\documentclass[preprint,12pt]{elsarticle}

\usepackage{tikz}
\usepackage{xcolor}
\usepackage{amssymb} 
\usepackage{amsmath}
\usepackage{hyperref}
\usepackage{booktabs}
\usepackage{multirow}
\usepackage[T1]{fontenc}
\usepackage{lipsum}

\begin{document}




\title{Inductive Graph Embeddings through Locality Encodings}

\author[1,2]{Nurudin Alvarez-Gonzalez\corref{cor1}}
\ead{nuralgon@gmail.com}

\author[2]{Andreas Kaltenbrunner}
\ead{andreas.kaltenbrunner@upf.edu}
\author[2]{Vicen\c{c} G\'omez}
\ead{vicen.gomez@upf.edu}

\cortext[cor1]{Corresponding author}

\address[1]{NTENT Hispania\\Avinguda Diagonal, 210\\08018 Barcelona, Spain}
\address[2]{Department of Information and Communications Technologies\\Universitat Pompeu Fabra\\T\`anger, 122-140. 08018 Barcelona, Spain}

\begin{abstract}
Learning embeddings from large-scale networks is an open challenge. Despite the overwhelming number of existing methods, is is unclear how to exploit network structure in a way that generalizes easily to unseen nodes, edges or graphs. In this work, we look at the problem of finding inductive network embeddings in large networks without domain-dependent node/edge attributes. We propose to use a set of basic predefined local encodings as the basis of a learning algorithm. In particular, we consider the degree frequencies at different distances from a node, which can be computed efficiently for relatively short distances and a large number of nodes. Interestingly, the resulting embeddings generalize well across unseen or distant regions in the network, both in unsupervised settings, when combined with language model learning, as well as in supervised tasks, when used as additional features in a neural network. Despite its simplicity, this method achieves state-of-the-art performance in tasks such as role detection, link prediction and node classification, and represents an inductive network embedding method directly applicable to large unattributed networks.
\end{abstract}

\begin{highlights}
\item We present a novel method of inductive graph representation learning that is directly applicable to networks without edge or node attributes based on local encodings of node neighbourhoods.
\item Our work shows that structural relationships within a graph can be represented in a compact embedding space, and that the resulting embedding vectors are effective in supervised and unsupervised learning tasks.
\item The proposed local encodings generalize to unseen graphs from the same domain in a multi-label node classification setting, outperforming methods based on Graph Convolutional Networks.
\item Our experiments show that execution times scale polynomially as the number of nodes grow in randomly generated graphs under equivalent generation parameters.
\end{highlights}


\maketitle              

\section{Introduction}
\label{sec:Introduction}
Graph structures are the natural way to represent arbitrary relationships in complex domains. Because of their generality, they can be applied to problems in which represented entities display rich interactions between each other. However, compactly capturing such interactions is a non-trivial task, often made unfeasible, for example, by the high-dimensionality and complexity of real-world networks.

Representation learning has been successfully applied to graphs, capturing useful interactions in compact ways~\cite{hamilton2017representation}. The learned representations can be used for downstream machine learning tasks, and have seen applications as varied as molecule generation for drug design~\cite{JMLR:v21:19-671}, content recommendation in social platforms~\cite{Ying-2018-PinSAGE}, or social network analysis~\cite{DBLP:journals/corr/PerozziAS14-deepwalk}.

A wide array of learning approaches have focused on local node features and interactions. \emph{Transductive} approaches require all nodes in the graph to be available at training time to learn a representation for each of them. Transductive methods can capture rich relationships within networks in a scalable manner. However, the resulting representations do not generalize to unseen nodes or edges, and require additional attention to deal with label permutations. 

In contrast, \emph{inductive} approaches~\cite{DBLP:journals/corr/HamiltonYL17-GraphSAGE} aim to compactly represent network interactions so that the specific identity of a node is irrelevant to its representation. Inductive representations are typically derived from attributes of nodes and edges within their neighbourhoods. In either case, inductive models tend to be more demanding than their transductive counterparts, particularly when entire graphs or node neighbourhoods ought to be sampled during training~\cite{DBLP:journals/corr/KipfW16-SemiSup-GCN}.  

In this paper, we introduce \textbf{IGEL} (\textbf{I}nductive \textbf{G}raph \textbf{E}mbeddings through \textbf{L}ocality Encoding), a novel inductive method for learning structural graph embeddings. IGEL aims to overcome the aforementioned issues in previous works by representing network structures explicitly, in the form of neighbourhood structure dictionaries. Dense vector representations are then learned from the structural dictionaries, in either an unsupervised graph-dependent process or under full supervision for a target task. 

We organize the paper as follows: in Section~\ref{sec:RelatedWork}, we discuss the main lines of ongoing research in the state-of-the-art. Afterwards, we describe our model in Section~\ref{sec:IGELOverview}. In Section~\ref{sec:Experiments} we present experimental results which consider different learning settings and an empirical anslysis of the complexity of IGEL. Finally, we discuss our results and conclude in Section~\ref{sec:Conclusions}.


\section{Related Work}
\label{sec:RelatedWork}
In this section, we briefly review the contributions most related to our work, grouping them according to their setting and their associated limitations. We then describe IGEL in this context, and give an introduction to how the method serves as a solution to existing problems. For a more comprehensive review of graph embedding techniques at large, we refer the reader to existing surveys~\cite{GOYAL201878, DBLP:journals/corr/abs-1901-00596}. 

\subsection{Unattributed Graph Embeddings}

First approaches to learning graph representations were applicable in unattributed settings as exemplified by models such as DeepWalk \cite{DBLP:journals/corr/PerozziAS14-deepwalk}. DeepWalk reduces the problem of learning compact node representations to the task of training a word2vec \cite{DBLP:journals/corr/abs-1301-3781} skip-gram language model over sequences of node labels sampled with random walks generated from a graph. Later, node2vec \cite{DBLP:journals/corr/GroverL16-node2vec} further extended this idea by introducing additional parameters to control the locality of the random walks, and thus, the properties of the learned latent spaces. 

An alternative transductive approach, LINE \cite{DBLP:journals/corr/TangQWZYM15-LINE}, tackled the problem by learning representations on the basis of structure. LINE minimizes the KL-divergence between the probability distribution of distinguishing adjacent nodes and between nodes with similar edges out of learned representations. The use of pairwise similarity measures has been further explored more recently by VERSE~\cite{DBLP:journals/corr/abs-1803-04742-VERSE}. VERSE tailors similarity measures such as personalized PageRank or SymRank into the loss function, aiming to capture explicit properties from underlying network relationships in contrast to context-based language models.

\textbf{Problem A.} The aforementioned methods are scalable and showed remarkable results in representing graphs in varied settings. However, they are limited by their nature as transductive approaches, namely, that the representations do not generalize to unseen nodes, edges or graphs.

\subsection{Convolutional Graph Networks}

More recently, research has focused on methods that learn inductive graph representations. Inductive representations are attractive as they can be applied on either transductive or inductive settings, so long as the same attributes are available on seen and unseen problem instances. Early models, such as Planetoid~\cite{DBLP:journals/corr/YangCS16-Planetoid} derived their representations from an embedding layer in a neural network whose input included node attributes, to perform semi-supervised training on classification tasks. 

However, a key component of graph data is that connected nodes can hold profound relationships between each other. Capturing those relationships can be a complex task, and much work has been devoted to it, particularly on works with Graph Convolutional Networks (GCNs) \cite{DBLP:journals/corr/NiepertAK16-GCN}. GCNs aggregate features from local neighbourhoods and the attributes of their nodes in order to generate a new feature representation for a given node. Because of this, they are ideal to capture the interactions between nodes and their neighbours. As such, GCNs have been successfully applied tasks such as semi-supervised learning ~\cite{DBLP:journals/corr/KipfW16-SemiSup-GCN} or node classification \cite{velikovi2017graph-GAT}. However, training GCNs when working on massive graphs is a challenge. 

In this direction, some work has focused on sampling the graph during training, as proposed by LGCL~\cite{DBLP:journals/corr/abs-1808-03965-LGCL} and GraphSAGE~\cite{DBLP:journals/corr/HamiltonYL17-GraphSAGE}. LGCL starts from a set nodes selected randomly and adds nodes incrementally by sampling according to a BFS strategy, until a budget of nodes is reached. Similarly, GraphSAGE represents a node by sampling a fixed number of neighbours at each distance from the node. This sampling approach bounds computational cost to a constant factor, trading off computational cost for predictive performance.

Based on this idea, methods such as PinSAGE~\cite{Ying-2018-PinSAGE} have shown promising results by building inductive models that can cope with web-scale problems by adapting sampling-based models to distributed environments.

\textbf{Problem B.} GCN models are applied in the presence of attributed data on nodes or edges. Applying them to unattributed networks is not straightforward, and it is not clear how the representations capture structural aspects or in tasks that require learning from node-to-node relationships. 

\subsection{Structural Representations}

To this end, models such as struct2vec~\cite{DBLP:journals/corr/FigueiredoRS17-struct2vec}, have been proposed to deal with graph substructures directly. Particularly, struct2vec tackles the issue by learning skip-gram representations on a constructed meta-graph that connects structurally-similar nodes within a threshold. 

\textbf{Problem C.} Constructing a meta-graph is a computationally costly process, as it relies on computing a pair-wise structural similarity measure across all nodes. Struct2vec proposes a series of optimizations to reduce the would-be quadratic computational cost, however, the learned representations remain label-dependent and, thus, they are affected by the aforementioned limitations of transductive approaches.

\subsection{Our Contribution in Context}

In this work, we tackle some of the problems we described about existing graph representation methods, namely:
\begin{enumerate}
    \item[\textbf{Problem A}] \emph{The difficulty in translating transductive network embeddings such as DeepWalk or node2vec to unseen graphs from the same domain}. Rather than working with individual nodes, we learn embeddings of graph structures, which capture more general properties of the graph.
    \item[\textbf{Problem B}] \emph{Graph Convolutional Networks on unattributed graphs leverage node and edge attributes}. We propose an inductive yet unattributed representation learning method. The resulting embeddings can serve potentially as additional input features to deep neural networks or graph convolutions.
    \item[\textbf{Problem C}] \emph{The scalability problems of existing structural embedding methods}. Methods such as struct2vec~\cite{DBLP:journals/corr/FigueiredoRS17-struct2vec} require to compute a similarity multilayer graph, which is expensive. We propose a similar structural representation, but avoid computing meta-graph structures by borrowing random-walk sampling techniques to learn an inductive representation.
\end{enumerate}

In the next section, we detail how IGEL functions, relating the building blocks of the method with the problems we seek to solve.

\section{The IGEL Algorithm} 
\label{sec:IGELOverview}
In this section we describe IGEL in detail. We first provide an overview of the method and then drill down into the two steps that compose its high level algorithm. As IGEL embeddings can be trained in either supervised or unsupervised settings, we close the section with a formal description of the different objectives, paving the way for the experimental section of this work.

\subsection{Overview}

The main idea of IGEL is to represent local graph structures inductively by learning dense embeddings over a sparse space of structural attributes that appear in the surroundings of nodes in a graph. These dense embeddings can be learned in supervised or unsupervised ways, with the overall process encompasing two main steps. For a given graph $\mathcal{G} = \langle V, E \rangle$ composed of a set of vertices $V$ and edges $E$, representing a node requires:

\begin{enumerate}
  \item \textbf{Encoding.} The first step encodes local graph structures in the form of a sparse vector recording the frequencies of structural attributes found for every given node $n \in V$.
  \item \textbf{Embedding.} The sparse encoding is used as input to an embedding model, that transforms the input vector into a dense representation of the aggregated structural attributes.
\end{enumerate}

\subsection{Encoding Local Graph Structures}
\label{sec:encoding}

We capture the local structure by considering the neighbourhood of a node. For a given node $n \in V$ and encoding distance $\alpha\in \mathbb{N}$, let $\mathcal{G}_{\alpha}(n)$ denote the sub-graph of $\mathcal{G}$ containing all nodes found at distance at most $\alpha$ from~$n$, and the edges connecting them. From the neighbourhood graph $\mathcal{G}_{\alpha}(n)$, we represent the structural features of node $n$ by considering the number of times a degree $d$ is found at distance $c$ in the graph $\mathcal{G}_\alpha(n)$, $c=0, \hdots, \alpha$.

\begin{figure}[!t]
    \centering
    \includegraphics[width=0.65\linewidth]{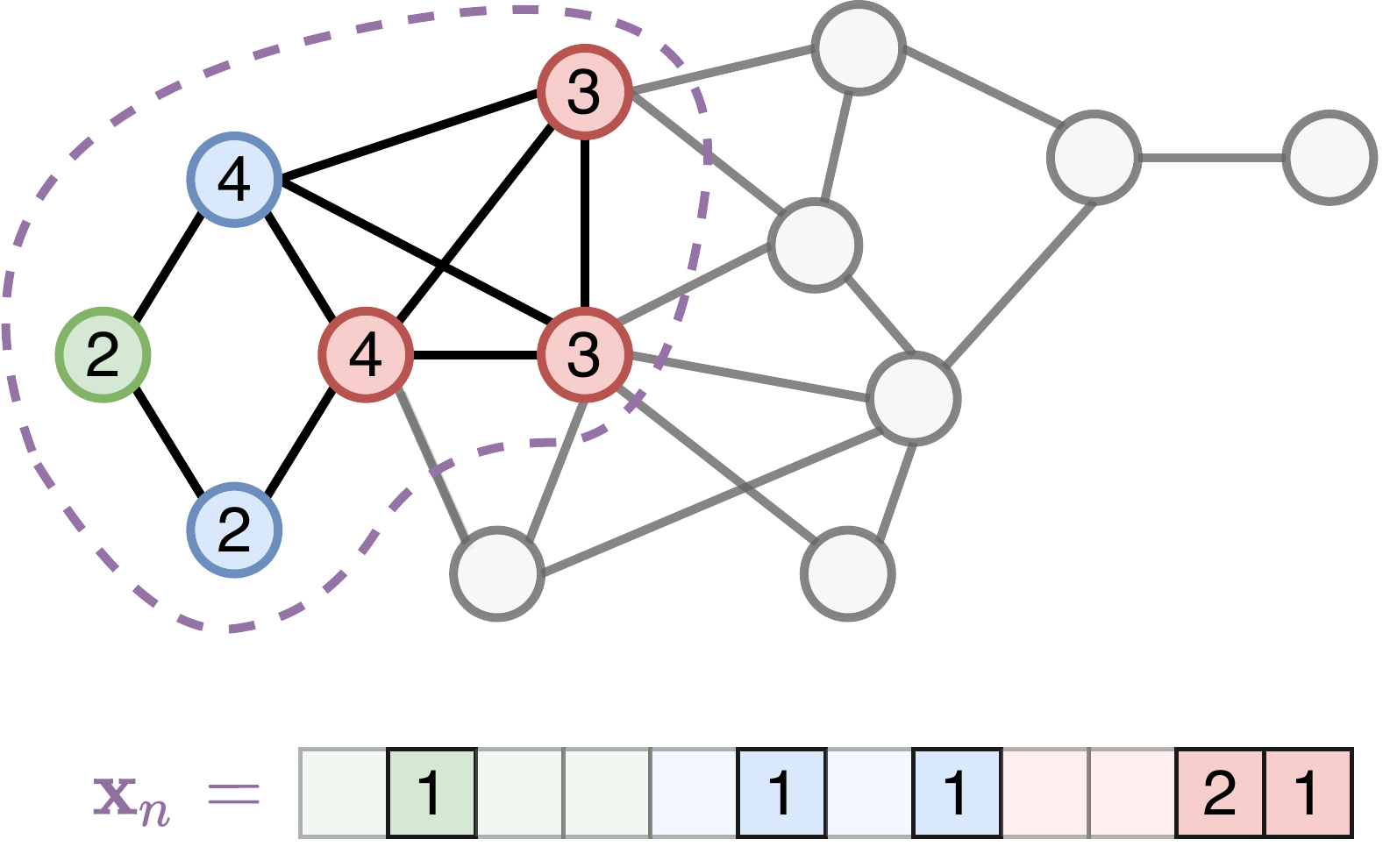}
    \caption{Encoding example. The dashed region denotes the corresponding neighbourhood of the green node $n$ for distance $\alpha = 2$. \textcolor{green!60!black}{Green nodes} are at distance \textcolor{green!60!black}{$0$}, \textcolor{blue!75!black}{blue nodes} at \textcolor{blue!75!black}{$1$} and \textcolor{red!80!black}{red nodes} at \textcolor{red!80!black}{$2$}. Numbers indicate the degrees within the induced graph. The structural representation for the green node is the following frequency mapping of distance-degree $(c, \delta)$ pairs: $\{
        \textcolor{green!60!black}{(0, 2): 1},
        \textcolor{blue!75!black}{(1, 2): 1},
        \textcolor{blue!75!black}{(1, 4): 1},
        \textcolor{red!80!black}{(2, 3): 2},
        \textcolor{red!80!black}{(2, 4): 1} 
    \}$, which translates into the sparse structural feature vector $\mathbf{x}_n$ shown at the bottom. 
    }
    \label{fig:GraphStructureEncoding}
\end{figure}

In this way, the structural features can be understood as a succession of distance $c \in \mathbb{N}$ and degree $\delta \in \mathbb{N}$ tuples $(c, \delta)$ and their respective counts. Given the max-degree $\delta_\text{max}$ of $\mathcal{G}$, a sparse feature vector $\mathbf{x}_n \in \mathbb{R}^\ell$, with $\ell =~\delta_\text{max} \times \alpha$, can be constructed such that the $i$th component of $\mathbf{x}_n$, $i = c\delta_\text{max}  + \delta$, contains the number of times the tuple $(c, \delta)$ appears in $\mathcal{G}_\alpha(n)$. A visual representation of this encoding process is shown in \autoref{fig:GraphStructureEncoding}.

The proposed encoding can be further processed and normalized. In this work, we log-normalize each feature count $z$ as $\log_2(1 + z)$ and then normalize the transformed features to the $[0, 1]$ interval. Also, if during testing we need to compute the features for a new node with degree $\delta'>\delta_\text{max}$, we use the bin corresponding $\delta_\text{max}$ to store its frequency.

\subsection{Embedding of the Sparse Representations}

\begin{figure}[!t]
    \centering
    \includegraphics[width=1.0\linewidth]{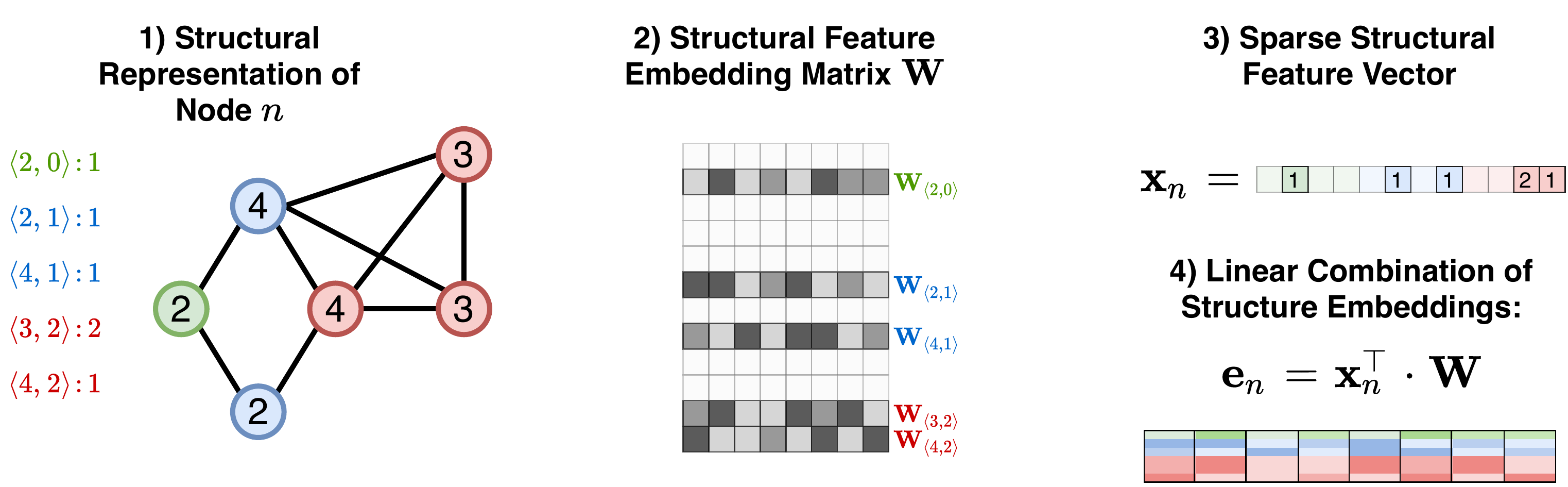}
    \caption{Structural embedding example. Shaded regions in $\mathbf{x}_n$ refer to the zero values in the sparse representation, which in turn represent unseen structures as seen in $\mathbf{W}$. The dense representation of node $n$ is $\mathbf{e}_n$, as seen in \autoref{eq:SimpleModel}. The final vector, represented in step $4$, is the linear combination of the structural feature matrix combined with the sparse feature vector, with each strip representing the value of the source structure in the embedding matrix and its width indicating the contribution to that particular component.}
    \label{fig:StructuralEmbedding}
\end{figure}

At this point, we can represent a node in terms of its immediate surroundings, but this representation is sparse and high dimensional. A simple way to transform the sparse structural features into a dense, lower dimensional, representation is through a linear transformation:
\begin{align}
    \mathbf{e}_n & =\mathbf{x}_n^\top \cdot \mathbf{W},
    \label{eq:SimpleModel}
\end{align}
where $\mathbf{e}_n$ is $d$-dimensional, and $\mathbf{W} \in \mathbb{R}^{\ell \times d}$ is a learnable matrix of parameters, $d\ll \ell$. Each row in $\mathbf{W}$ captures the vector representation of each single structural feature. Intuitively, this can be understood as a weighted average of the encoded structural features, representing the node in terms of its immediate surroundings. \autoref{fig:StructuralEmbedding} provides a visual overview. 

\subsection{Learning the Embeddings}

We now describe the learning procedure for the parameters $\mathbf{W}$ such that the general structural properties of the original encoding are preserved.Such properties to preserve are task-dependent: 
in exploratory network analysis, we aim to learn representations that capture general graph properties such as degree distributions, centrality measures, or structural similarities. In classification tasks, the objective is to identify attributes that are useful to discriminate between the classes, if the model can capture them. We now define the two optimization objectives considered by IGEL corresponding to both scenarios.

\subsubsection{Unsupervised Setting}

Our objective in the unsupervised setting is to capture the global network features in terms of the local structures surrounding every node. We formulate this problem similarly to skip-gram based methods~\cite{DBLP:journals/corr/GroverL16-node2vec,DBLP:journals/corr/PerozziAS14-deepwalk}, which optimize a likelihood function from sample data generated through random walks on the graph. In the following, we describe the two steps involved in learning the embeddings. 

\begin{figure}[!t]
    \centering
    \includegraphics[width=0.65\linewidth]{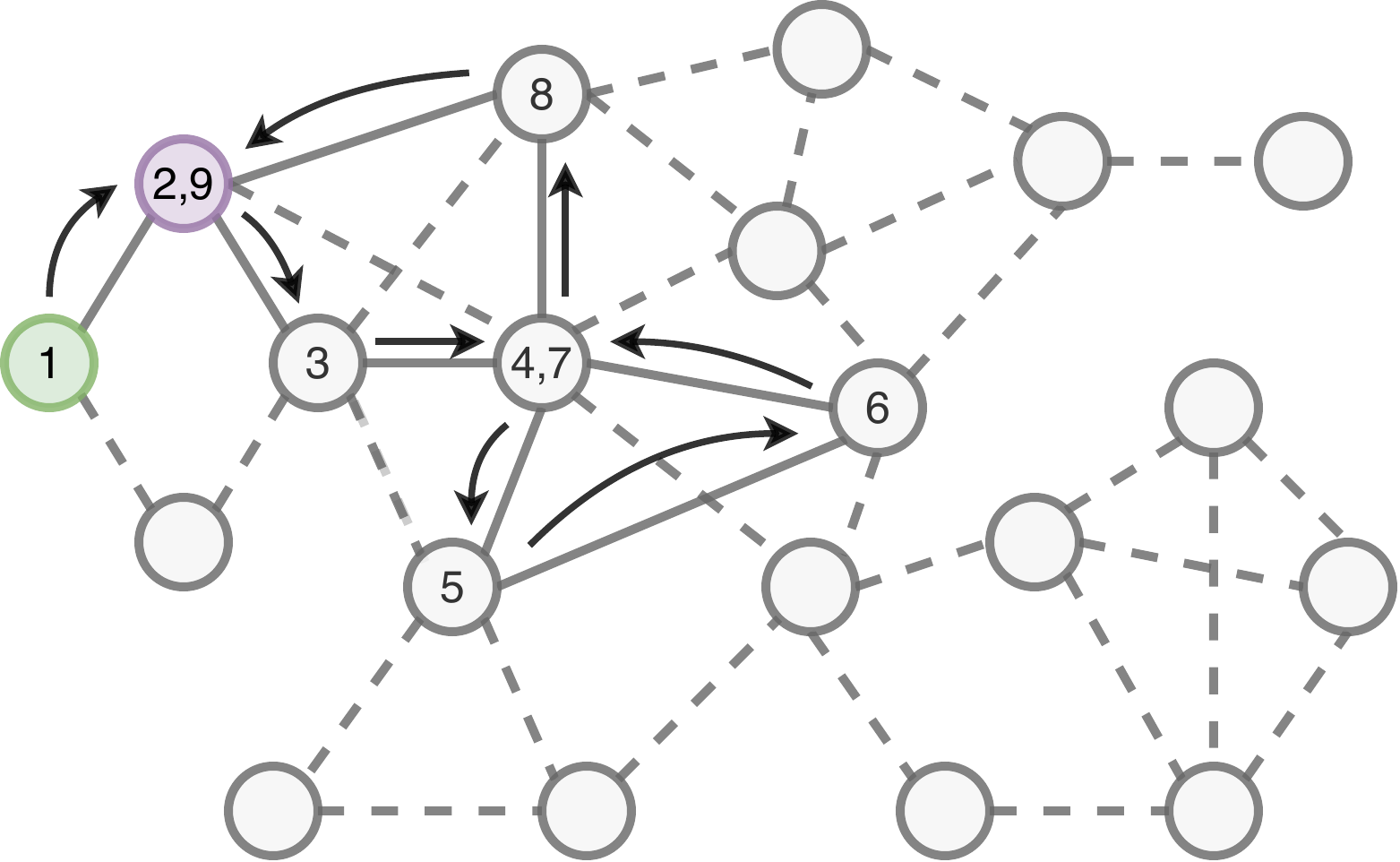}
    \caption{Example of random walk. Nodes contain the time-step when they were visited. The context of nodes seen besides a target node across the walk serves to learn the representations given by the approach shown in \autoref{fig:GraphStructureEncoding}.}
    \label{fig:GraphStructureSampling}
\end{figure}

\emph{Distributional Sampling through Random Walks}---
Our sampling strategy generates sample paths over the graph, selecting the next node uniformly from the neighbours of the current node. \autoref{fig:GraphStructureSampling} illustrates a possible random walk of length 9 in the same graph of~\autoref{fig:GraphStructureEncoding}. Each node contains the time-step(s) when it was visited during the random walk.

This step is similar to existing sample-based transductive methods, such as DeepWalk~\cite{DBLP:journals/corr/PerozziAS14-deepwalk}. In contrast to other methods such as node2vec~\cite{DBLP:journals/corr/GroverL16-node2vec}, our method uses unbiased random walks, and does not require tuning of the probabilities of in-out and return paths when sampling each graph. Additionally, our approach does not require any previous computation of a structural meta-graph, as proposed by struct2vec~\cite{DBLP:journals/corr/FigueiredoRS17-struct2vec}. Instead, we perform $w$ random walks of length $s$ starting from each node in the network.

As we randomly sample the structure of the graph, we produce sequences of traversed nodes. These sequences are used in the optimization objective we describe next.

\emph{Negative Sampling Optimization Objective}---
Given a random walk $\omega$, defined as a sequence of nodes of length $s$, we define the context $\mathcal{C}(n, \omega)$ associated to the occurrence of a node $n$ in $\omega$ as the the sub-sequence in $\omega$ containing the nodes that appear close to $n$, including repetitions. Closeness is determined by the hyper-parameter $p$, the size of the positive context window, i.e., the context contains all nodes that appear at most $p$ steps before/after the node within $\omega$.

Similarly to DeepWalk or node2vec, we define a skip-gram negative sampling objective that learns to identify nodes appearing in similar contexts within the random walks. Given a node $n_t \in V$ in random walk $\omega$, our task is to learn embeddings that 
assign high probability for nodes $n_o\in V$ appearing in the context $\mathcal{C}(n_t, \omega)$, and lower probability for nodes not appearing in the context. As we focus on the learned representation capturing these symmetric relationships, the probability of $n_o$ being in $\mathcal{C}(n_t, \omega)$ is given by:
\begin{align}
 p\left(n_o \in \mathcal{C}(n_t, \omega)\right) &= \sigma(\mathbf{e}^\top_{n_t} \cdot \mathbf{e}_{n_o}),
\end{align}
where $\sigma(\cdot)$ is the logistic function. 
\begin{table}[!t]
\centering
\caption{Hyper-Parameters used in unsupervised tasks.}
\begin{tabular}{@{}ll@{}}
\toprule
\textbf{Hyper-Parameter}    & \textbf{Description}                 \\ \midrule
$w \in \mathbb{N}$ & Number of random walks per node       \\ 
$s \in \mathbb{N}$ & Number of steps in the Random Walks  \\ 
$z \in \mathbb{N}$ & Proportion of negative examples      \\ 
$p \in \mathbb{N}$ & Positive samples context window size \\ 
\midrule \end{tabular}
\label{tab:UnsupervisedIGELParams}
\end{table}

The global objective function is the following negative sampling log-likelihood to be maximised. For each random walk and each node in the random walk, we sum the term corresponding to the positive cases for structures found in the context, and the expectation over $z$ negative, randomly sampled, structures:
\begin{align}\label{eq:negsamp}
  \mathcal{L}_{\text{u}}(\mathbf{W})&= \sum^{w}_{j=1} \sum_{
  \substack{n_t \in \omega_j,\\
  n_o \in \mathcal{C}(n_t, \omega_j)}}  \Bigg[\log \sigma(\mathbf{e}_{n_t}^\top \cdot \mathbf{e}_{n_o}) + \sum^z_{i=1}  \mathbb{E}_{{n_i} \sim P_n(V)} \Big[\log \sigma(-\mathbf{e}_{n_t}^\top \cdot \mathbf{e}_{n_i})\Big]\Bigg],
\end{align}
where $P_n(V)$ is the noise distribution from which the $z$ negative samples are drawn.
We maximize~\eqref{eq:negsamp} through gradient ascent. \autoref{tab:UnsupervisedIGELParams} summarizes the hyper-parameters in the unsupervised setting.

\subsubsection{Supervised Setting}

Besides its direct usage as an inductive node embedding method, IGEL can also be used as a node embedding layer in a deep neural network. In particular, the embedding of Eq.~\eqref{eq:SimpleModel} can be fed into a downstream model as additional input features for a multilayer perceptron (MLP) or a graph neural network (GNN). 
In this case, the weight matrix $\mathbf{W}$ is jointly learned with the rest of the model parameters $\boldsymbol{\Theta}$. Instead of capturing global structural properties through random walks as before, the model in the supervised setting captures salient structural features that have predictive power in a particular task.

\begin{figure}[!t]
    \centering
    \includegraphics[width=0.7\linewidth]{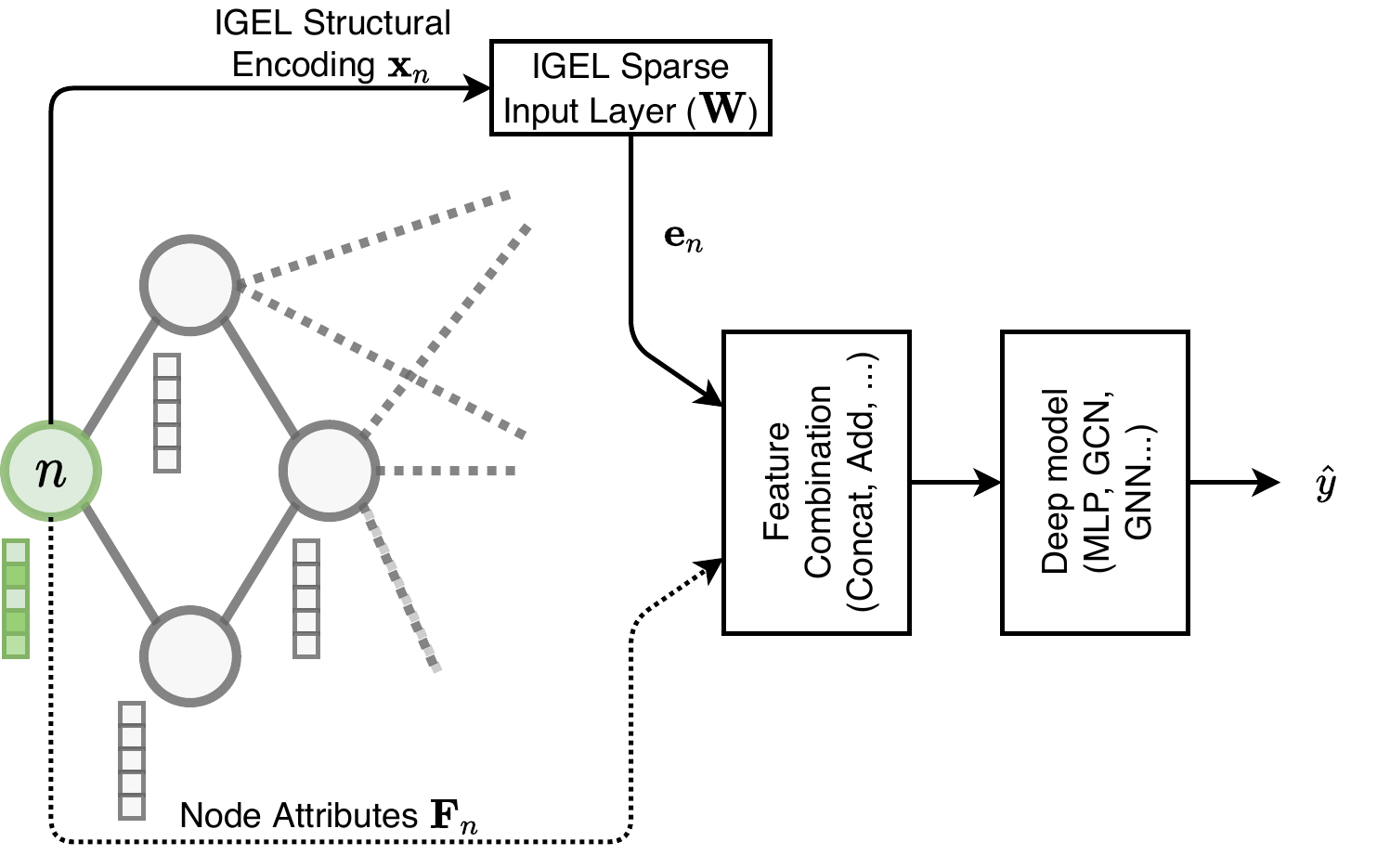}
    \caption{An example of IGEL applied in a supervised setting.
    The node embedding $\mathbf{e}_n$ provides a structural representation that can optionally be combined with additional node or edge features as an input to a deep learning model which outputs $\tilde{y}$.
    The weight matrix $\mathbf{W}$ is jointly learned with the rest of the parameters $\boldsymbol{\Theta}$.
    See text for details.}
    \label{fig:GraphSupervisedLearning}
\end{figure}

\autoref{fig:GraphSupervisedLearning} illustrates the general model for a standard node classification task, in which the IGEL embeddings are used in combination with possibly additional node attributes $\mathbf{F}_n$ to generate a final output $\hat{y}$ that is used to predict the node class label.

Using IGEL as a feature layer additionally allows for pre-training and fine-tuning when data is not sufficiently available. This is similar to natural language domains, where language models are first trained on large corpora and then retrained for specific tasks at hand.
Analogously, the IGEL layer can be learned first in the unsupervised manner described in the previous section, and subsequently refined to a specific inference task. 

Without loss of generality, we focus on multi-label classification without pre-training. For $M$ independent labels, the task is to minimise the binary cross-entropy of the predicted labels associated to nodes:
\begin{align}\label{eq:supervised}
\mathcal{L}_{\text{s}}(\mathbf{W},\boldsymbol{\Theta}) &= - \sum_{n\in V}\sum_{i=1}^M \Big[ y_n^i
\log
\sigma(\hat{y}_n^i) +
(1-y_n^i) \log\left(1-\sigma(\hat{y}_n^i)\right)\Big],
\end{align}
where $y_n^i$ is one if $i$ is the true class label of node $n$, and zero otherwise, and $\hat{y}_n^i$ is the network output corresponding to the class with label $i$.

\subsection{Computational Complexity}
We now describe the computational complexity of IGEL in relation to other embedding methods. Time and parameter complexities of other methods is described as in~\cite{GOYAL201878, DBLP:journals/corr/abs-1901-00596}. We differentiate time complexity, expressed in terms of computational steps given an input graph, and parameter complexity, which refers to the number of learnable parameters corresponding to a particular model. We note that certain parameters such as the number of random walks or their length in sampling-based methods appear as a constant, which we do not include in our analysis. 

We first consider the parameter complexity. Let $\gamma$ be the number of bins required to encode the local degree-based representation, which we set to the maximum degree $\delta_{\text{max}}$ in this work\footnote{For highly skewed degree distributions, the original algorithm can be modified to require less parameters $\gamma \ll \delta_{\text{max}}$ by using log-sized bins or exploiting the sparsity of the feature vectors.}. Since each embedding vector has dimension~$d$ and we have $\gamma$ bins for each encoding distance, the number of parameters scales as $\mathcal{O}(\alpha\times\gamma\times d)=\mathcal{O}(\ell\times d)$.

On the other hand, the time complexity involves two main steps: a pre-processing step that computes the encoding of the local representations, and the computations required at each optimization step. The encoding step requires to visit all nodes at distance $\alpha$ from each node $n$, and repeat that for each node in the graph, therefore the time complexity is exponential in~$\alpha$, $\mathcal{O}(|V|\times \delta_{\text{max}}^{\alpha})$. Once the local representations are computed, however, the optimizations step can be done efficiently, involving only a product that grows linearly with the number of parameters, or a forward-pass in a neural network.

\autoref{tab:ComplexityComparison} compares the time and parameter complexities of several models. Transductive models such as DeepWalk, node2vec or struct2vec scale linearly with respect to the number of nodes, both in terms of time and parameters. Inductive methods based on  graph convolutions incur in higher training costs, as signals have to be propagated across the whole set of nodes and edges. Their parameter complexity, however, is not directly dependent on either, and is instead a function of the number of input features. Sampling-based methods such as GraphSAGE reduce the time complexity cost by ensuring that a fixed number of sampled nodes are used during training and inference.

We note that IGEL lies at the crossroads of the previous methods. Its time complexity depends on both the number of nodes and the size of their surrounding neighbourhoods, similar to sampling-based GCNs. However, IGEL lacks the fixed bounds in the number of operations that is guaranteed by sampling a constant number of nodes at a time. Due to this fact, we expect IGEL's performance to degrade on denser graphs with neighbourhoods containing a wider range of different degrees among their nodes. Finally, IGEL's parameter complexity is closer to transductive methods, with a bound given by the maximum degree in the graph.

\begin{table}[!h]
\centering
\caption{A comparison of graph representation methods in terms of their average parameter and time complexities. The parameters $\alpha$ stands for IGEL's encoding distance, $r$ is the sampling factor for sampling GCNs, $\rho$ is the convolutional distance in GCNs, $\delta_{\text{max}}$ is the maximum degree and $d$ is either the size of the embedding vectors or the number of features, depending on the method.}
\begin{tabular}{@{}lrr@{}}
\toprule
\textbf{Name}                    & \textbf{Time Complexity}           & \textbf{Parameter Complexity} \\ \midrule
\textit{DeepWalk, node2vec}      & $\mathcal{O}(|V|)$                 & $\mathcal{O}(|V|d)$           \\
\textit{struct2vec\textsuperscript{*}} & $\mathcal{O}({|V|} \log |V|)$      & $\mathcal{O}(|V|d)$      \\
\textit{GCN}                     & $\mathcal{O}(\rho |V| |E|)$        & $\mathcal{O}(\rho d^2)$       \\
\textit{GraphSAGE}               & $\mathcal{O}(r^\rho |V|)$          & $\mathcal{O}(\rho d^2)$       \\
\textit{IGEL}                    & $\mathcal{O}(\delta_{\text{max}}^\alpha |V|)$   & $\mathcal{O}(\alpha \delta_{\text{max}} d)$     \\ \bottomrule
\end{tabular}
\label{tab:ComplexityComparison}
\caption*{\scriptsize \textsuperscript{} \textsuperscript{*}Struct2vec's authors describe its complexity in two terms. First, an $\mathcal{O}({|V|} \log |V|)$ preprocessing step to generate a structural meta-graph. Second, the structural embedding training process, equivalent to running DeepWalk over the meta-graph. We report the dominating time complexity of the preprocessing step.} 
\end{table}

\section{Experimental Evaluation}
\label{sec:Experiments}
In this section, we evaluate IGEL in different tasks. We first analyze visually the type of structural properties captured by the method. Then, we evaluate the method in link prediction tasks for medium-sized graphs. Finally, we apply IGEL to an inductive multilabel node classification task in a large collection of protein graphs. For this analysis, we use the graphs described in \autoref{tab:GraphsOverview}.
\begin{table}[!t]
\centering
\caption{Overview of the graphs used in the experiments.}
\begin{tabular}{@{}lrrr@{}}
\toprule
                                                             & ~\textbf{\# Nodes} & ~\textbf{\# Edges}  & ~\textbf{Avg. Clust. Coeff.} \\ \midrule
Les Miserables (cloned)                                      & 154       & 509     & 0.73 \\
Sister Cities~\cite{AKaltenbrunner13-SisterCities}           & 11~618    & 15~225  & 0.11 \\
Facebook~\cite{snapnets}                                     & 4~039     & 88~234  & 0.52 \\
ArXiv AstroPhysics~\cite{snapnets}                           & 18~772    & 198~050 & 0.32 \\
PPI (train)~\cite{DBLP:journals/corr/HamiltonYL17-GraphSAGE} & 44~906    & 613~184 & 0.12 \\ \bottomrule
\end{tabular}
\label{tab:GraphsOverview}
\end{table}
The goal of this section is to evaluate experimentally the following hypothesis:
\begin{enumerate}
    \item \textit{Generality and scalability:} the method should be capable of representing structural aspects in small and large graphs alike. 
    \item \textit{Inductivity:} the method should learn representations from graphs seen during training that generalize to unseen graphs from the same domain.
    \item \textit{Scalability and compatibility with deep learning approaches:}
    the method should outperform or be competitive with existing deep neural methods when used in conjunction with them without incurring in too excessive additional costs.
\end{enumerate}

\subsection{Unsupervised Setting}
We first evaluate IGEL in the unsupervised setting using networks that are amenable for visual analysis. We focus on analyzing how the encoding distance parameter~$\alpha$ influences the learned representations.

\begin{figure}[!b]
    \includegraphics[width=.49\textwidth]{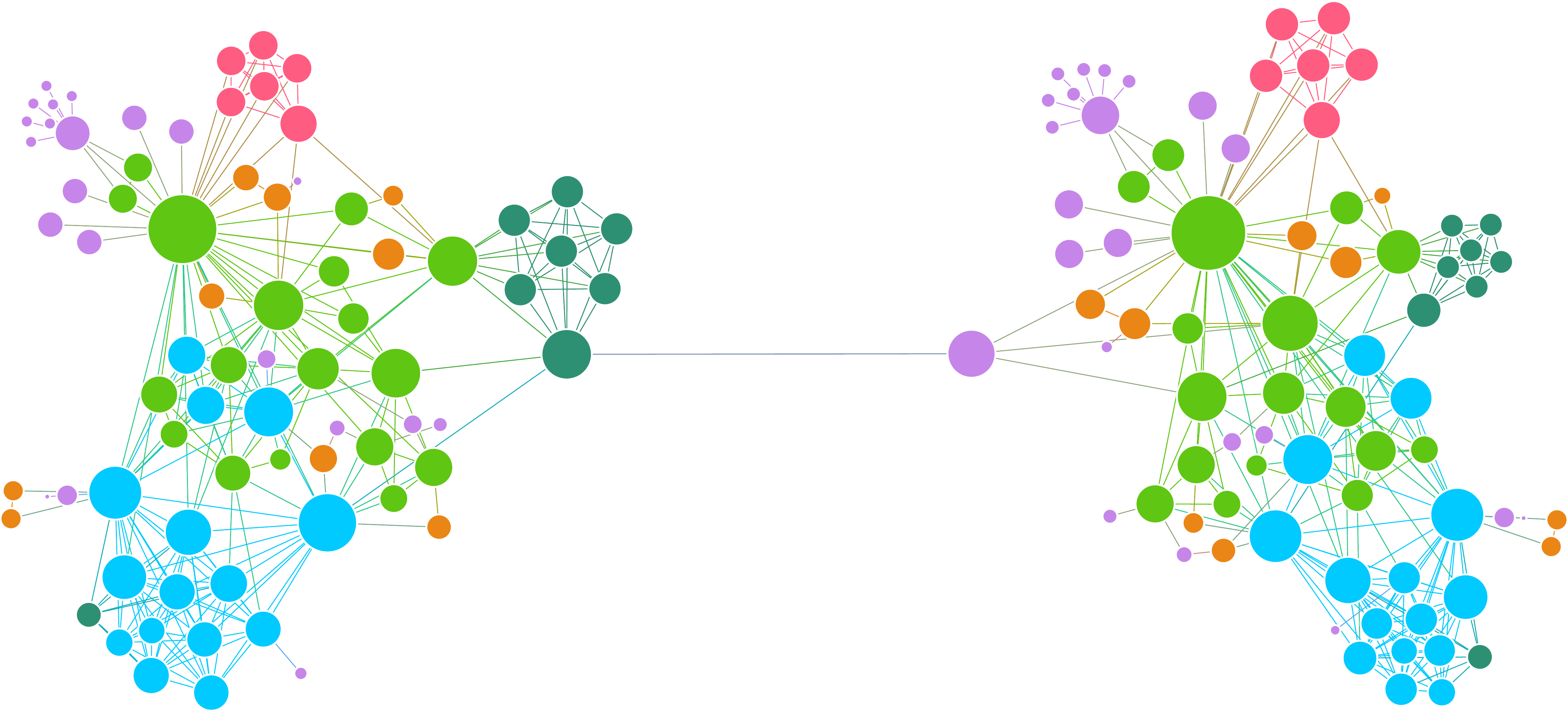}
    \includegraphics[width=.49\textwidth]{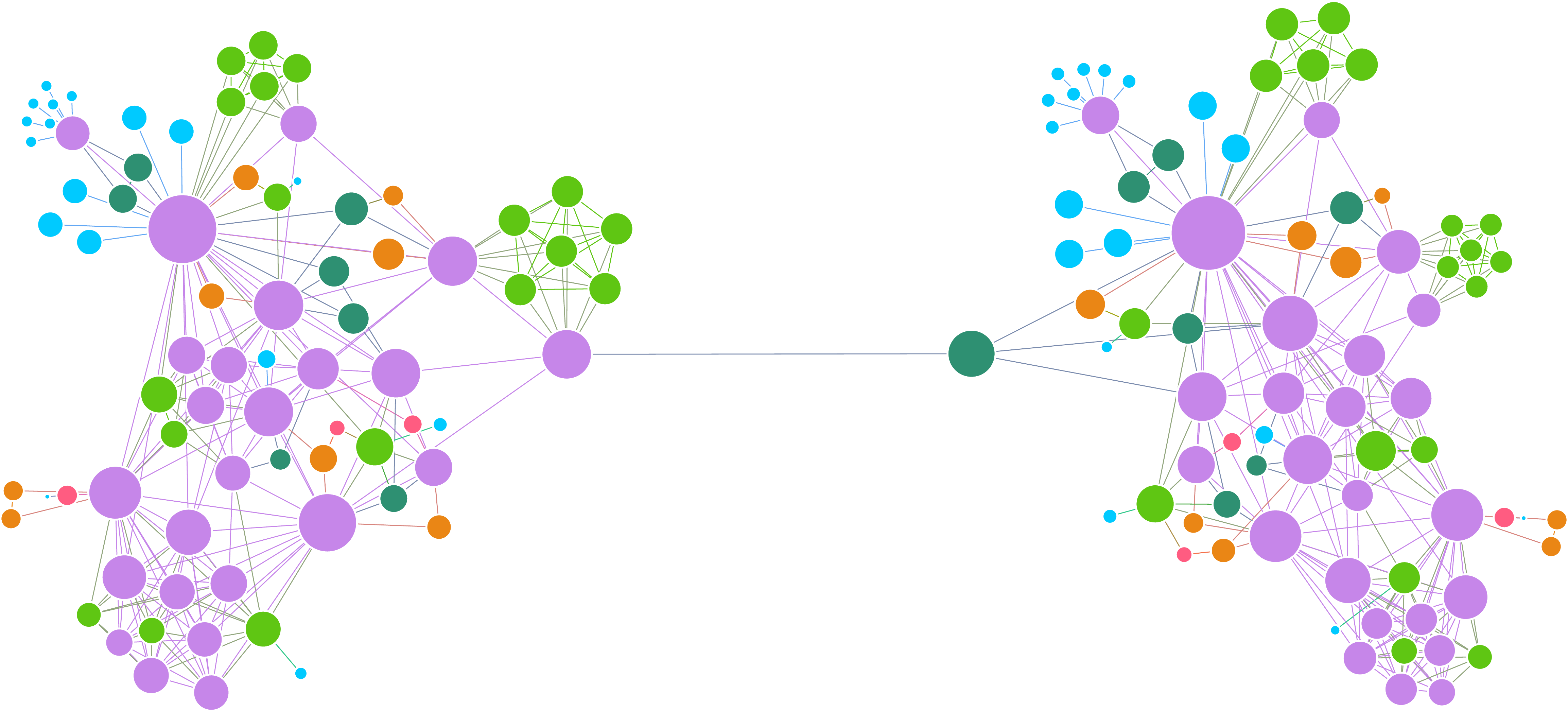}
    \caption{Visualization of the embeddings in the (cloned) Les Mis\'erables network: node colors reflect the clustering of the embeddings obtained for two different encoding distances, $\alpha = 1$ (left) or $\alpha = 2$ (right).
    In addition to capture community structure, the IGEL embeddings also generalize structurally similar nodes between distant regions.
    Node sizes correspond to their harmonic closeness centrality score. }
    \label{fig:MiserablesClustering}
\end{figure}

\emph{Les Misérables Network}--- 
In this task, we consider a variation of the original Les Misérables network~\cite{miserables} in which two identical copies of the network are connected by an edge between two nodes from each copy selected randomly, as shown in~\autoref{fig:MiserablesClustering}. This allows us to analyze how the IGEL embeddings can capture structural similarity from distant areas of the graph.

We first use IGEL to learn feature representations for every node, and subsequently cluster the obtained embeddings. For a complete description of the experiment and parameter settings, see~\ref{app:case}.

\autoref{fig:MiserablesClustering} shows the composite network with nodes colored based on their clusters. For $\alpha = 1$, the local encodings only consider direct neighbours. In this case, the resulting embeddings capture the standard notion of community or homophily, i.e., nodes with the same color are densely connected between each other and weakly connected to the rest. Furthermore, the embeddings corresponding to nodes within the two copies of each community also appear clustered together, although some nodes are not directly connected. This shows that IGEL embeddings can capture 
structural similarity, not only locally, but also from distant nodes in the graph.
Larger encoding distances allow IGEL to capture relationships on larger portions of the graph. For $\alpha = 2$, shown in~\autoref{fig:MiserablesClustering}(b),
the embeddings are clustered in two major groups, separating frequently visited nodes---with higher degrees and closeness centrality, and hence more commonly appearing in the sampling process---from connecting bridges or edges.

\emph{Sister Cities Network}--- 
We now illustrate how IGEL can be used for knowledge discovery using a larger network: the network of Sister Cities~\cite{AKaltenbrunner13-SisterCities}, a graph mined from Wikipedia where nodes correspond to cities and links to partnerships. We follow a similar approach as before and train IGEL embeddings for each node. We then perform a clustering analysis of the embeddings. \ref{app:sister} shows the details of the setup.

\begin{figure}[!t]
    \centering
    \includegraphics[width=\textwidth]{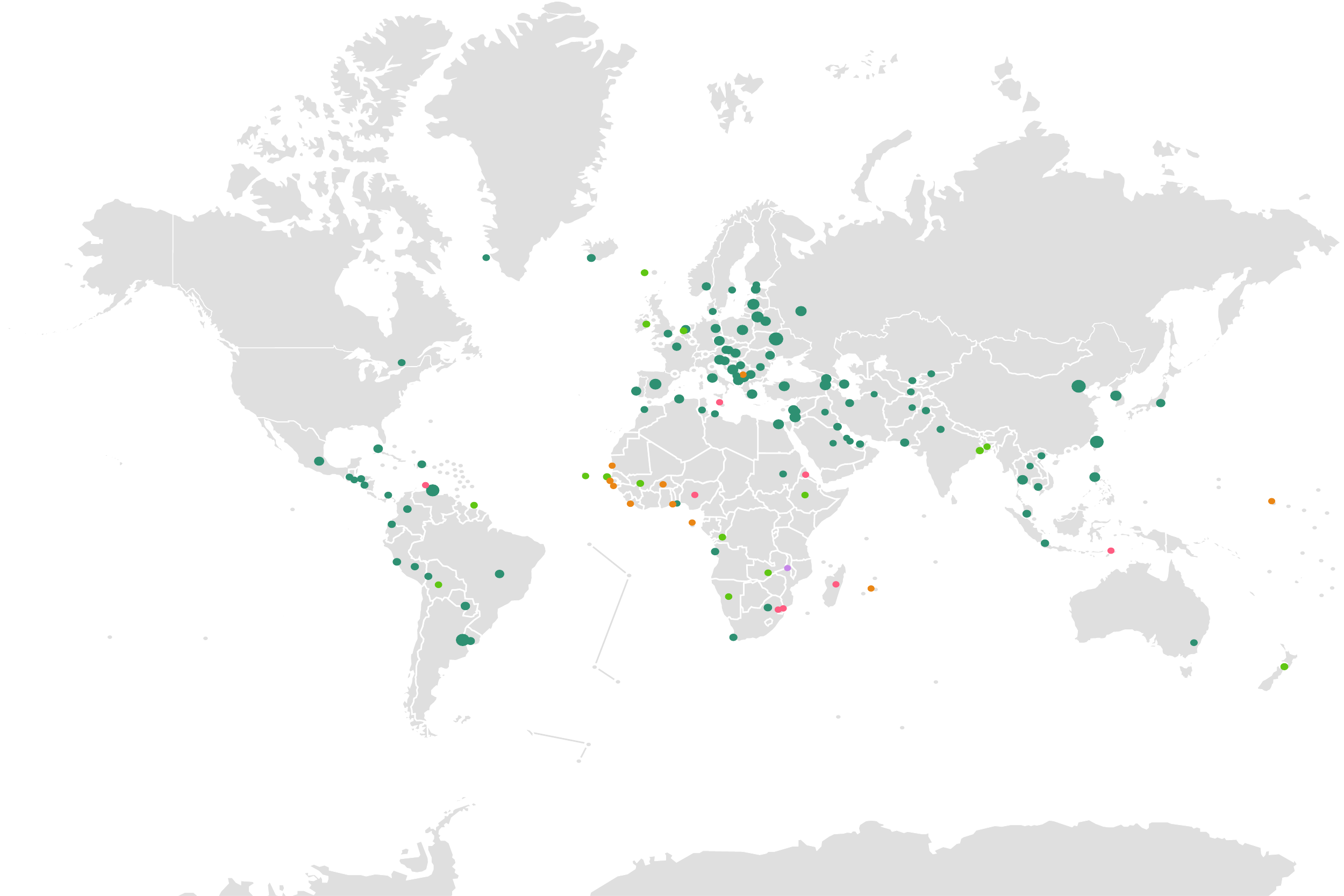}
    \caption{Geo-plot of the learned clusters on the Sister Cities dataset, filtering only cities that are capitals. Each point is a city, with a radius proportional to its degree, and colored by the group assigned by the clustering model. To learn the clusters, we use an IGEL encoding distance $\alpha = 4$ and rum KMeans with $k = 6$. }
    \label{fig:SisterCities}
\end{figure}
The obtained clusterings allows us to identify interpretable roles for the nodes, for example, cities that are country capitals. To quantitatively test this, we use an external labeled dataset that assigns a total of $179$ nodes as country capitals, and then evaluate IGEL as a binary classifier by treating the nodes that belong to the cluster containing the highest number of capital cities as positive, and the rest of the nodes as negative.

\autoref{fig:SisterCities} shows a geo-plot with label colors corresponding to a subset of $872$ capital cities. Remarkably, this cluster composed of IGEL embeddings consistently outperforms heuristic methods by capturing up to 28\% more country capitals, as measured by F1-score (see~\autoref{tab:SisterCitiesF1Scores} in the Appendix for details). From these results, we can conclude that IGEL is also effective discovering non-trivial embeddings that are interpretable.

\subsection{Link Prediction}
We now evaluate IGEL on the task of predicting whether an edge exists between two given nodes. To generate training data, we follow the methodology proposed in~\cite{DBLP:journals/corr/GroverL16-node2vec}, which allows us to compare with previous transductive algorithms.

In particular, for a given graph, we generate negative examples (non-existing edges) by sampling random pairs of nodes that are not connected in the original graph. Positive examples (existing edges) are obtained by removing half of the edges while keeping the mutilated graph after the edge removals connected\footnote{The constraint of keeping the graph connected is only required to compare with transductive methods and not necessary for IGEL.}. Both sets of node pairs are chosen to have the same cardinality.

The mutilated graph is used to train the IGEL embeddings using the unsupervised objective of Equation~\eqref{eq:negsamp}. Once the weight matrix~$\mathbf{W}$ has been learned, we formulate a classification task, to classify pairs of nodes according to the labels described before.

For this final step, we use a logistic regression classifier with features corresponding to the Hadamard (element-wise) product of the embedding vectors of the nodes at each end of the predicted edge. This is the best combination of features reported in~\cite{DBLP:journals/corr/GroverL16-node2vec}. Note that this setting can be seen as a special case of our Equation~\eqref{eq:supervised} with $M=2$ and the deep architecture being replaced by a single layer of additional weights.

We optimize all hyper-parameters described in~\autoref{tab:UnsupervisedIGELParams}, together with the encoding distance, using grid search. Our experimental setting involves link prediction on in a social media setting in the Facebook ego-network graph, and a citation network on arXiv capturing collaboration between scientists. We observe that the performance only depends strongly in two parameters: the encoding distance $\alpha$ and, to a lesser extent, the number of negative samples $z$. In general, we observe a diminishing returns effect, e.g., for $\alpha=3$ the performance is comparable to that of using $\alpha = 2$ while taking an order of magnitude more time. In Section~\ref{sec:Scalability} we evaluate empirically the impact of $\alpha$ in the complexity of the algorithm.

\begin{table}[ht]
\centering
\caption{Area Under the ROC Curve (AUC) results for the Facebook and AstroPhysics arXiv graphs. IGEL outperforms transductive methods, only using local degree structure. Hyper-parameters are optimized using grid search in IGEL, and transductive methods are the best ones reported in~\cite{DBLP:journals/corr/GroverL16-node2vec}.}
\begin{tabular}{@{}lrr@{}}
\toprule
\textbf{Method}                                                                     & \textbf{Facebook}          & \textbf{arXiv}           \\ \midrule
DeepWalk \cite{DBLP:journals/corr/PerozziAS14-deepwalk} & 0.968            & 0.934          \\
LINE \cite{DBLP:journals/corr/TangQWZYM15-LINE}         & 0.949            & 0.890          \\
node2vec \cite{DBLP:journals/corr/GroverL16-node2vec}   & 0.968            & 0.937          \\ \midrule
IGEL                                                     & \textbf{0.976}   & \textbf{0.984}  \\\bottomrule
\end{tabular}
\label{tab:LinkPredictionAUROCResults}
\end{table}

\autoref{tab:LinkPredictionAUROCResults} shows the best performing configuration on a series of five independent experiments. The best performing configuration on the Facebook graph features $\alpha = 2$, learning $d = 256$ component vectors with $e = 10$ walks per node, each of length $s = 150$ and $p = 8$ negative samples per positive for the unsupervised negative sampling. Respectively on arXiv, we find the best configuration at $\alpha = 2$, $d = 256$, $e = 2$, $s = 100$ and $p = 9$. Finally, we note that we run each experiment on a machine with 8 cores, 16GB of memory and a single GPU with 8GB of memory. 

Compared to previous transductive approaches, which represent every individual node independently, our method represents nodes in terms of the aggregated representations of structural features within their neighbourhoods, which can lead to the same representation shared by different nodes. Despite this, IGEL outperforms previous methods such as DeepWalk, LINE and node2vec, that have access to the whole graph when learning node representations.

\subsection{Node Classification}
In this section, we evaluate how IGEL is able capture structural attributes that generalize to unseen graphs from the same domain. For that, we consider an inductive multi-label node classification task, in which we explore the performance of IGEL with or without pre-training and fine-tuning. The objective is to predict $121$ different binary labels capturing protein functions in a series of Protein-to-Protein Interaction (PPI) graphs. Aside from the graph structures, every node has an additional set of attributes, allowing us to evaluate how the performance of IGEL differs in attributed networks. In this case, we use the supervised setting with the objective of Equation~\eqref{eq:supervised}.

We use the supervised baselines proposed by GraphSAGE~\cite{DBLP:journals/corr/HamiltonYL17-GraphSAGE} and LCGL~\cite{DBLP:journals/corr/abs-1808-03965-LGCL}, and further compare with a fully supervised version of GraphSAGE~\cite{velikovi2017graph-GAT}. We perform randomized grid search to identify the best possible set of parameters for an MLP containing IGEL and (optionally) feature inputs.

We analyze both encoding distances $\alpha \in \{1, 2\}$, with other IGEL hyper-parameters being fixed after a small greedy search based on the best configurations in the link prediction experiments. For the MLP model, we perform a greedy architecture search, including the number of hidden units, activation functions and depth. Our results show scores averaged over five different seeded runs with the same configuration, as obtained from the parameter search. We run each experiment on a machine with 8 cores, 16GB of memory and a single GPU with 8GB of device memory. 

\begin{table}[!ht]
\centering
\caption{Micro-F1 scores for the multilabel classication task on the PPI graphs. IGEL combined with node features used as input for an MLP outperforms LGCL and supervised GraphSAGE both with and without sampling.}

\begin{tabular}{@{}llr@{}}
\toprule
\multicolumn{2}{l}{\textbf{Method}}             & \multicolumn{1}{l}{\textbf{PPI}} \\ \midrule
Random\textsuperscript{$\ddagger$}         &               & 0.396                            \\
Only Features (MLP, ours)                  &               & 0.558                            \\ \midrule
GraphSAGE-GCN\textsuperscript{$\dagger$}  &               & 0.500                             \\
GraphSAGE-mean\textsuperscript{$\dagger$} &               & 0.598                             \\
GraphSAGE-LSTM\textsuperscript{$\dagger$} &               & 0.612                             \\
GraphSAGE-pool\textsuperscript{$\dagger$} &               & 0.600                             \\ \midrule
GraphSAGE (no sampling)\textsuperscript{$\ddagger$} &     & 0.768                   \\
LGCL\textsuperscript{*}                   &               & 0.772                   \\ \midrule
\multirow{2}{*}{IGEL ($\alpha = 1$)}       & Graph Only    & 0.736                              \\
                                          & Graph + Feats & \textbf{0.850}                            \\ \midrule
\multirow{2}{*}{IGEL ($\alpha = 2$)}       & Graph Only    & 0.506                            \\
                                          & Graph + Feats & 0.741                            \\ \midrule
\end{tabular}
\caption*{\scriptsize \textsuperscript{} Results as reported by $\dagger$: \cite{DBLP:journals/corr/HamiltonYL17-GraphSAGE} $\ddagger$: \cite{velikovi2017graph-GAT} $*$: \cite{DBLP:journals/corr/abs-1808-03965-LGCL}}
\label{tab:PPIResults}
\end{table}

\autoref{tab:PPIResults} shows the classification results in terms of micro-averaged F1 score. Our model using structural embeddings and node features significantly outperforms both LGCL and every GraphSAGE instantiation with $\alpha = 1$. In this case, we find that a higher value of $\alpha = 2$ produces worse results than $\alpha=1$, which we attribute to overfitting. We believe that using regularization would help further improve these results. For reproducibility, we note that the best performing configuration is found with $\alpha = 2$ on $d = 256$ length embedding vectors, concatenated with node features as the input layer for 1000 epochs in a 3-layer MLP using ELU activations with a learning rate of 0.005. Additionally, we apply 100 epoch patience for early stopping monitoring the F1 score on the validation set.

Remarkably, we note that a model trained using only graph structural features at $\alpha = 1$ is able to outperform every GraphSAGE configuration that uses sampling. This highlights the strong predictive power of structural features alone, which is often disregarded when node or edges attributes are available.

\subsection{Complexity Analysis}\label{sec:Scalability}
We finish this experimental evaluation with an empirical analysis of the computational complexity of IGEL. For that, we evaluate our implementation in the unsupervised setting using synthetic data generated according to the Erdös-Renyi model. We control for three aspects that condition the runtime of the algorithm: the number of nodes $|V|$, the encoding distance $\alpha$, and the average degree per node. 

The training process is then performed with a realistic parameter configuration for encoding distances $\alpha= 1,2,3$. For each training configuration, we execute five independent experiments. Each run is performed on a Slurm-based cluster requesting 48 cpu cores and 200GB of memory on each setting. We record the overall training times, and analyze the run-time as a function of the size of the graph and the average degree.
\begin{figure}[!ht]
    \centering
    \includegraphics[width=.49\linewidth]{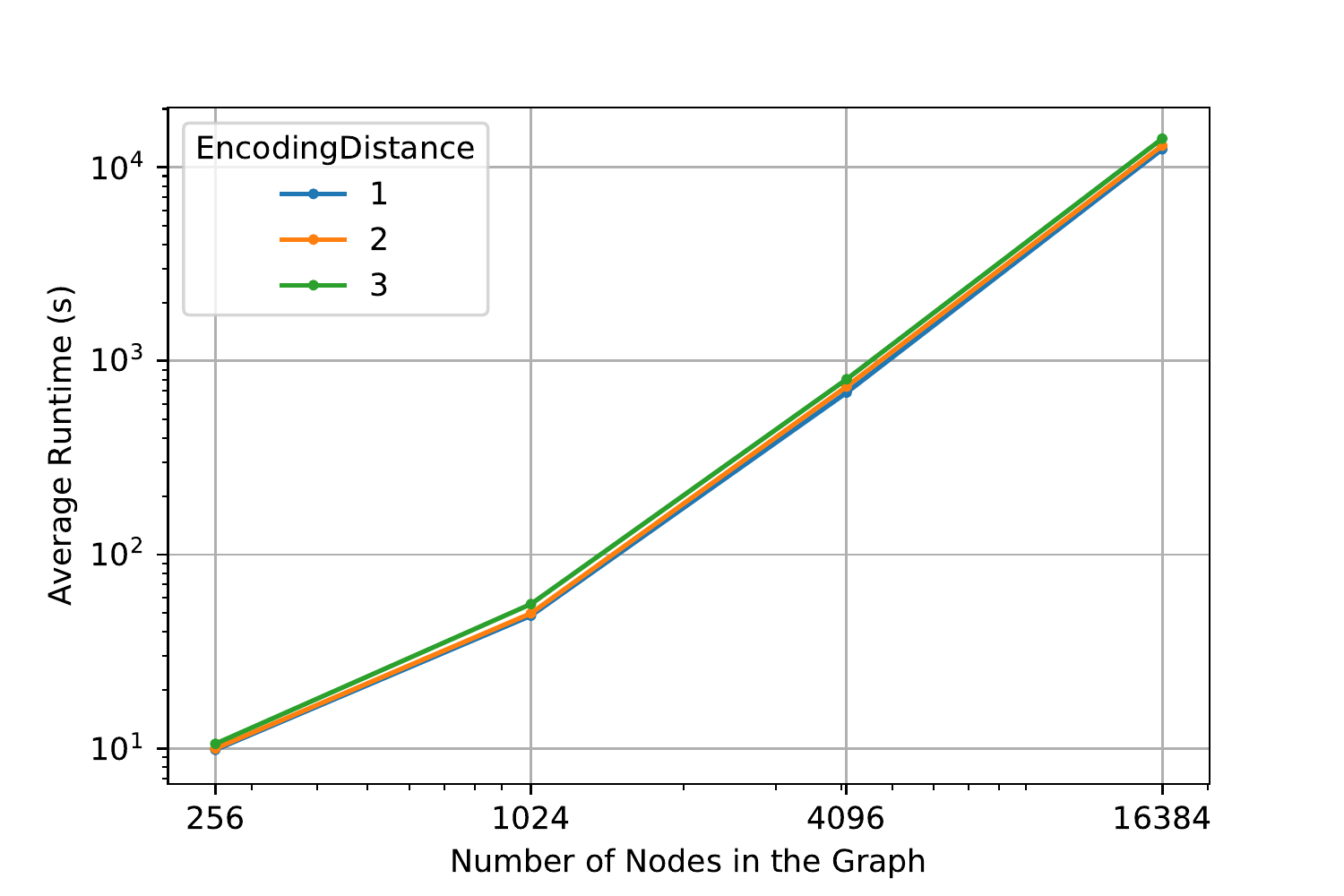}
     \includegraphics[width=.49\linewidth]{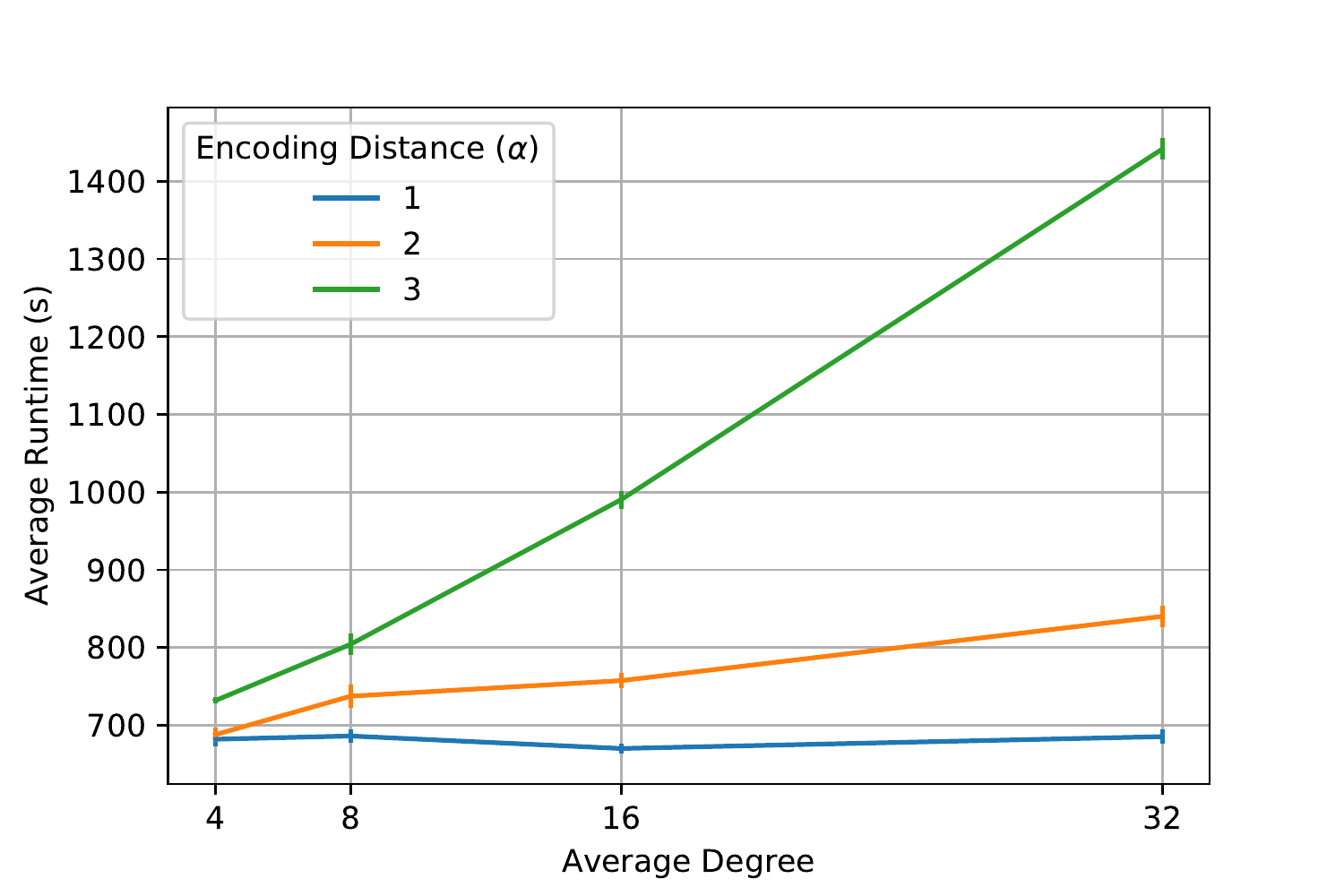}
    \caption{Scalability analysis.\textbf{(left)} Log-log plot showing the average runtime as a function of the graph size $|V|$ for different values of encoding distance $\alpha$ with fixed average degree of 8. \textbf{(right)} Average runtime as a function of the average degree for different values of $\alpha$ with fixed size of $|V|=4,096$.}
    \label{fig:ScalabilityResults}
\end{figure}

\autoref{fig:ScalabilityResults} (a) shows the run-time as a function of the number of nodes $|V|$ for different values of the encoding distance $\alpha$, keeping the average degree fixed.
We observe that $\alpha$ does not significantly affect the cost in this setting. Furthermore all curves show an approximate linear scaling for $|V|\geq 1,024$ (note the log-scale in both axes), showing that IGEL can be applied to large-scale graphs in this regime.

We also analyze the influence of the average degree for different values of $\alpha$ in~\autoref{fig:ScalabilityResults} (b).
In this case, for $\alpha = 1$, we observe an almost constant
runtime as a function of the average degree, suggesting that the direct structural features can be feasibly applied to denser graphs. On the other hand, the runtime grows linearly with the average degree for values of $\alpha>2$, which is an actual limitation of IGEL in large-scale, dense networks\footnote{Our Python code, available at \url{https://github.com/nur-ag/IGEL}, can be admittedly improved using a sparse implementation of the structural representations, or a more memory-efficient programming language.}.

Overall, these results show that IGEL is a scalable method for values of $\alpha = 1,2$. We believe that ideas from sampling-based GCN literature can be extended to purely structural methods like IGEL to allow for the encoding and learning of very distant structural features.

\section{Conclusions}
\label{sec:Conclusions}
In this work we have presented IGEL, a novel inductive network embedding method that compactly captures graph structures in a dense latent space. To the best of our knowledge, IGEL is the first inductive embedding algorithm explicitly designed for unattributed graphs that captures structural features without the need for node attributes. 

Our results show that IGEL is scalable and can compete with transductive approaches in tasks such as link prediction on unattributed graphs. Additionally, IGEL can inductively generalize to unseen graphs, outperforming models based on Graph Convolutional Networks in a multi-label classification setting. Our work also highlights future directions for research, namely, regarding the improvement of time and memory consumption, finding ways to capture global aspects of a graph that IGEL cannot represent, and managing the evolution of degree-based structural features in temporal graphs.



\section*{Acknowledgements}
Nurudin's research is partially funded by project 2018DI047 of the Catalan Industrial Doctorates Plan.
The project leading to these results has also received funding from ``La Caixa'' Foundation (ID 100010434), under the agreement LCF/PR/PR16 /51110009. 
Vicenç G\'omez is supported by the Ramon y Cajal program RYC-2015-18878 (AEI/MINEICO/FSE,UE).

\bibliographystyle{splncs04}
\bibliography{references}{}

\appendix
\section{Additional Results on Les Mis\'erables Network}\label{app:case}
We obtain our results on the Les Mis\'erables Network running unsupervised IGEL for 5 epochs with a batch size of $50~000$, an Adam learning rate of 0.01, and 200 steps per random walk. The positive context window size is 10, with a proportion of 10 negative samples per positive and 8-component embedding vectors. We direct the reader to the code repository, in which \texttt{configs/clustering.json} contains the exact hyperparameter configuration used.

\begin{figure}[!ht]
    \centering
    \begin{tikzpicture}
      \node (img)  {\includegraphics[width=0.8\linewidth]{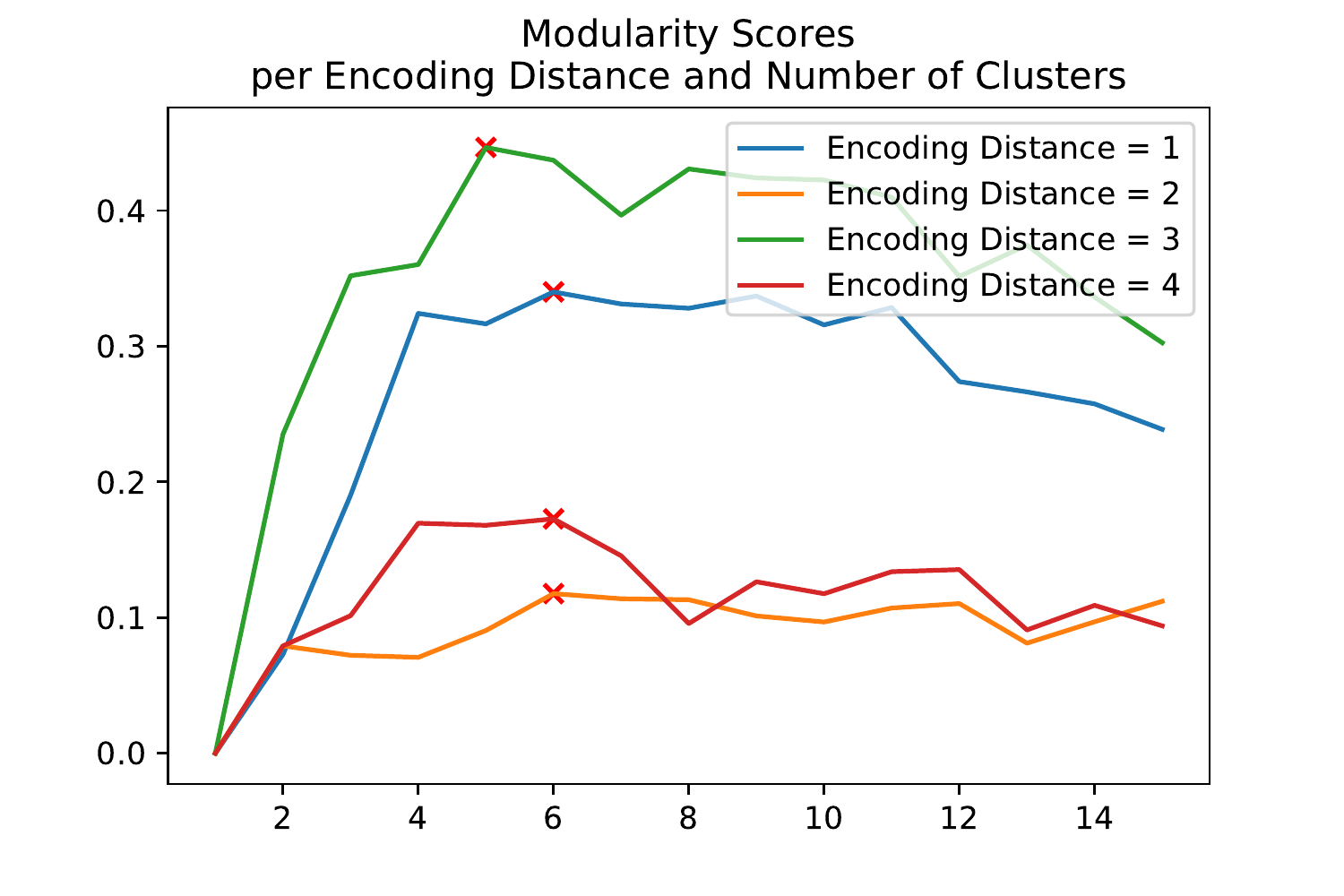}};
      \node[node distance=0cm, yshift=-3.6cm] {Number of clusters $k$};
      \node[node distance=0cm, rotate=90, anchor=center,yshift=5.1cm] {Modularity Score};
    \end{tikzpicture}
    \caption{Modularity score for the different encoding distance values when compared with the total number of clusters. The red markings represent the scores of the best performing total communities for each encoding distance value.}
    \label{fig:MiserablesModularity}
\end{figure}

To select the number of clusters used in \autoref{fig:MiserablesClustering}, we use modularity score~\cite{PhysRevE.69.026113-modularity}. \autoref{fig:MiserablesModularity} shows that highest modularity is obtained consistently with 5 or 6 clusters for different values of the encoding distance $\alpha$.

\begin{figure}[!ht]
    \centering
    \includegraphics[width=0.7\linewidth]{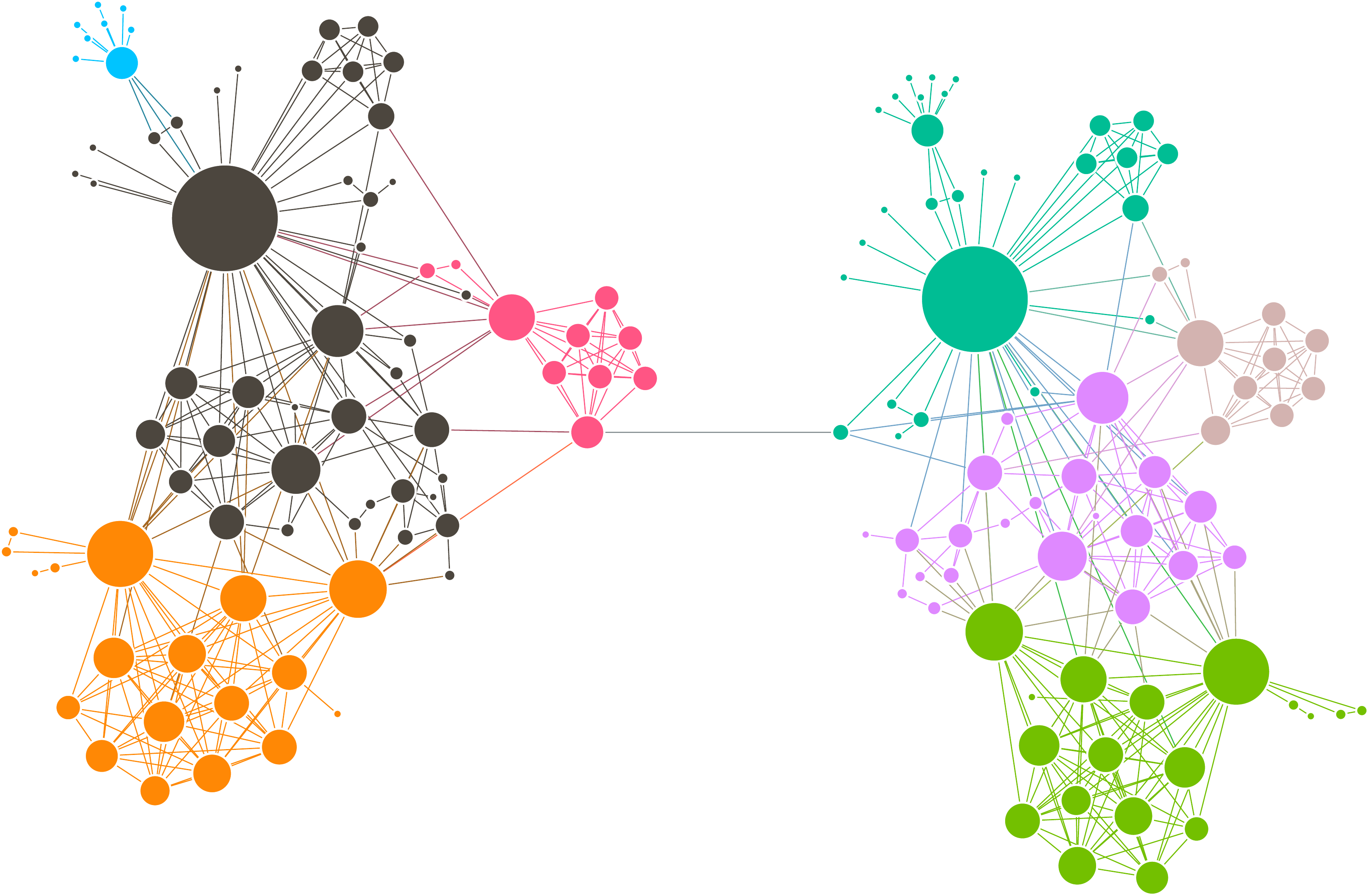}
    \caption{Node clusters produced by the Louvain Community Detection method. In contrast with IGEL, Louvain directly optimizes modularity over the whole graph. This noticeably makes the clusters across copies of the graph different, whereas IGEL can only identify the same representations given the available structural information in the neighbourhood of a node.}
    \label{fig:MiserablesLouvain}
\end{figure}

The result of running a standard community detection method in the graph is shown in \autoref{fig:MiserablesLouvain}, where we can observe that the local communities correspond to clustered embeddings found by IGEL for $\alpha=1$. However, IGEL is able to capture the additional structural similarity due to the cloned nature of the two subgraphs.

\section{Additional Results on Sister Cities Network}\label{app:sister}

We obtain our results on the Sister Cities Network running unsupervised IGEL for 2 epochs with a batch size of 25000, an Adam learning rate of 0.01, and 30 steps per random walk. The positive context window size is 7, with a proportion of 10 negative samples per positive and 30-component embedding vectors. We direct the reader to the code repository, in which \texttt{configs/clusteringLarge.json} contains the exact hyper-parameter configuration used.

\emph{Correlation with node indicators}---
For every node, we compute a score given only the embeddings of nodes in the graph, and then study its correlation to PageRank, Closeness, Betweenness and Degree centralities. To do so, we start with the intuition that a highly central node should be highly similar to its neighbours, since its representation should minimise the distance with any of the surrounding nodes. With this in mind, we compute the per-node scores as the sum of similarity scores between the node embedding and the embeddings of every one of its neighbours. Finally, our similarity function is the dot product between embedding vectors. 

Since we are comparing distributions that are defined over different domains, such as probabilities in the case of PageRank or integers in the case of degree, we focus on the rank rather than the actual values between scores and metrics. Thus, we use Spearman's $\rho$ correlation statistic to measure the monotonicity between both. The results for different encoding distances are shown in \autoref{tab:SisterCitiesCorrelations}.

\begin{table}[!ht]
    \caption{Spearman Correlations (with all $p < 10^{-20}$) between the node-to-graph self-similarity scores and graph centrality metrics.}
    \begin{tabular}{@{}crrrr@{}}
    \toprule
    \multicolumn{1}{l}{\textbf{Encoding Dist.}} & \multicolumn{1}{l}{\textbf{PageRank}} & \multicolumn{1}{l}{\textbf{Betweenness}} & \multicolumn{1}{l}{\textbf{Closeness}} & \multicolumn{1}{l}{\textbf{Degree}} \\ \midrule
    $\alpha = 1$                                        & 0.754                                 & 0.791                                    & 0.319                                  & 0.899                               \\
    $\alpha = 2$                                        & 0.740                                & 0.842                                   & 0.299                                 & 0.665                              \\
    $\alpha = 3$                                        & 0.772                                & 0.769                                   & 0.284                                 & 0.907                              \\
    $\alpha = 4$                                        & 0.776                                & 0.769                                   & 0.281                                 & 0.903                              \\
    \bottomrule
    \end{tabular}
    \label{tab:SisterCitiesCorrelations}
\end{table}

The correlation between embedding scores and results show several patterns. First, the embedding scores between a node and its neighbours is highly correlated with Degree, PageRank and Betweenness centralities. Second, closeness centrality is the least correlated of the metrics, as it measures how near a node is to every other node, in effect capturing a global property that needs to look beyond immediate node neighbourhoods. This falls in line with the hypothesis that we discussed in the previous section, in which larger graphs should display scores that are less correlated with global metrics. We direct the reader to \autoref{tab:MiserablesCorrelations} for the correlations in the Les Misérables graph, which exhibit higher correlations with the metrics across the board, and more so with closeness in particular.

\emph{Analysis of Capitals}---To compare the clustering model, we set a baseline with a heuristic approach based on node degree and PageRank score. We rank the nodes by either metric and take as many nodes as there are in the positive cluster. We then compute F1-scores for the cluster, degree and PageRank approaches, as shown in \autoref{tab:SisterCitiesF1Scores}. 

Upon qualitative inspection, we additionally notice that the clusters associated with capital cities usually include large, densely populated cities. Additionally, we found that the model often discriminates against capital cities located in Africa. We propose this may be caused by a reduced coverage of sister relationships in Wikipedia, or from the lack of such links between African cities and the rest of the world. 

\begin{table}[!ht]
\caption{Comparison between F1-scores of the cluster containing the most capitals $C$ and the result taking the top $|C|$ nodes as ranked by degree or PageRank. \textbf{Bold} means that the configuration is the best in a row, \textit{italic} that it is the best result for the encoding distance in that row. }
\centering
\begin{tabular}{@{}rr|rrrrr@{}}
\toprule
\multicolumn{1}{c}{\textbf{$\alpha$}} & \multicolumn{1}{c|}{\textbf{$k$}} & \multicolumn{1}{c}{\textbf{$|C|$}} & \multicolumn{1}{c}{\textbf{$C$ F1}} & \multicolumn{1}{c}{\textbf{\begin{tabular}[c]{@{}c@{}}Top-$|C|$\\ Degree F1\end{tabular}}} & \multicolumn{1}{c}{\textbf{\begin{tabular}[c]{@{}c@{}}Top-$|C|$\\ PageRank F1\end{tabular}}} & \multicolumn{1}{c}{\textbf{\begin{tabular}[c]{@{}c@{}}\%vs\\ Degree F1\end{tabular}}} \\ \midrule
\multirow{3}{*}{1}               & 5                                 & 854                                    & \textbf{0.197}                      & 0.174                                                                                    & 0.070                                                                                      & 13.32                                                                                 \\
                                 & 6                                 & 848                                    & \textbf{0.201}                      & 0.175                                                                                    & 0.070                                                                                      & 14.44                                                                                 \\
                                 & 7                                 & 839                                    & \textit{\textbf{0.202}}             & 0.177                                                                                    & 0.071                                                                                      & 14.42                                                                                 \\ \midrule
\multirow{3}{*}{2}               & 5                                 & 578                                    & \textbf{0.248}                      & 0.198                                                                                    & 0.095                                                                                      & 25.34                                                                                 \\
                                 & 6                                 & 566                                    & \textit{\textbf{0.258}}             & 0.201                                                                                    & 0.097                                                                                      & 28.02                                                                                 \\
                                 & 7                                 & 568                                    & \textbf{0.257}                      & 0.201                                                                                    & 0.096                                                                                      & 27.99                                                                                 \\ \midrule
\multirow{3}{*}{3}               & 5                                 & 1~892                                  & \textbf{0.118}                      & 0.116                                                                                    & 0.075                                                                                      & 1.73                                                                                  \\
                                 & 6                                 & 886                                    & \textit{\textbf{0.195}}             & 0.178                                                                                    & 0.068                                                                                      & 9.47                                                                                  \\
                                 & 7                                 & 1~360                                  & \textbf{0.144}                      & 0.140                                                                                    & 0.070                                                                                      & 2.78                                                                                  \\ \midrule
\multirow{3}{*}{4}               & 5                                 & 2~181                                  & \textbf{0.108}                      & 0.105                                                                                    & 0.074                                                                                      & 2.48                                                                                  \\
                                 & 6                                 & 872                                    & \textit{\textbf{0.196}}             & 0.173                                                                                    & 0.069                                                                                      & 13.23                                                                                 \\
                                 & 7                                 & 1~224                                  & 0.141                               & \textbf{0.151}                                                                           & 0.068                                                                                      & -6.62                                                                                 \\ \bottomrule
\end{tabular}
\label{tab:SisterCitiesF1Scores}
\end{table}

\begin{figure}[!t]
    \centering
     \includegraphics[width=\linewidth]{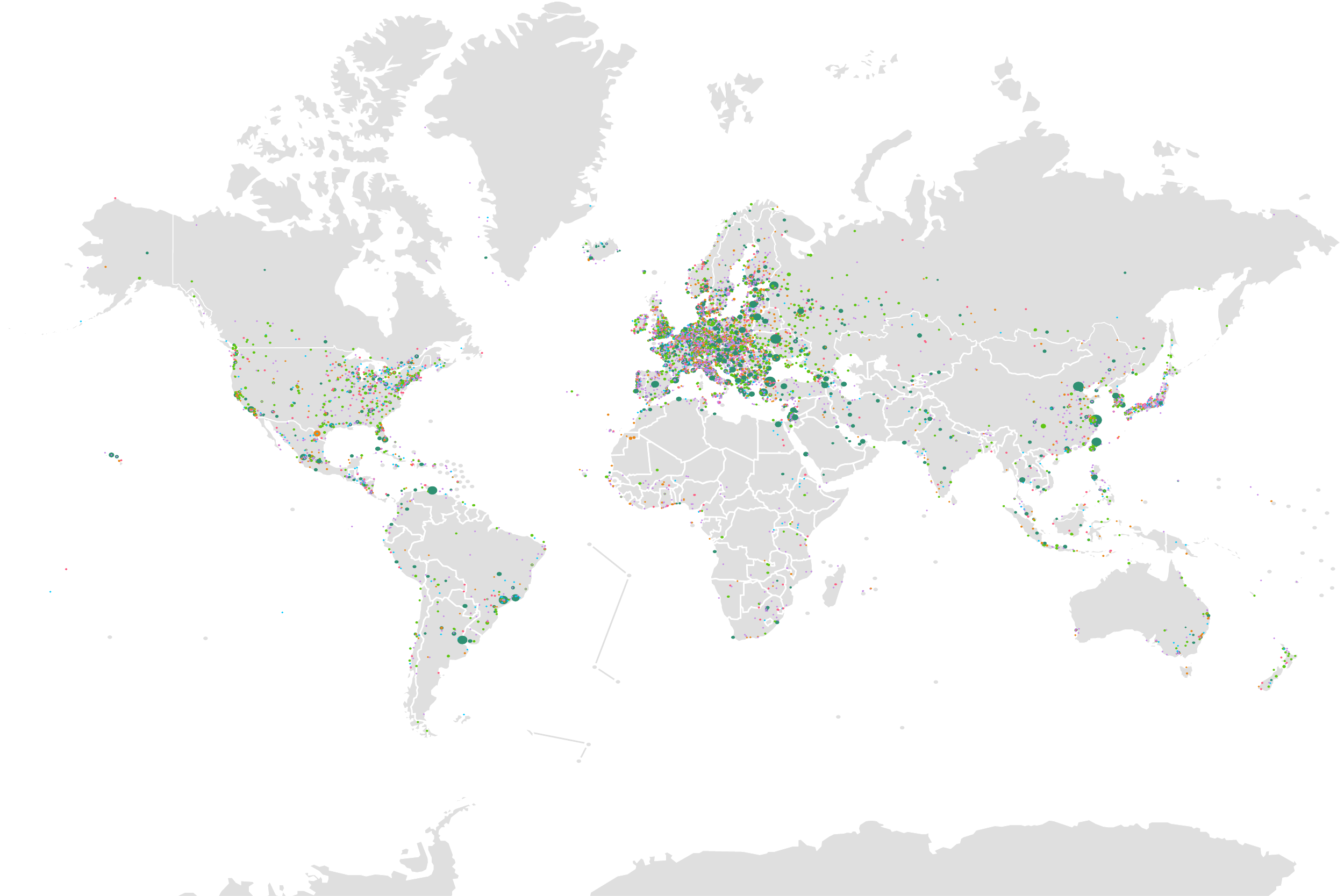}
    \caption{Clusters of Sister Cities found using IGEL embeddings with $\alpha = 4$ and $k = 6$.}
    \label{fig:SisterCitiesClusters}
\end{figure}

\begin{table}[!t]
    \caption{Spearman Correlations (with $p < 10^{-3}$) between the node-to-graph self-similarity scores and graph centrality metrics on the Les Miserables cloned graph.}
    \begin{tabular}{@{}crrrr@{}}
    \toprule
    \multicolumn{1}{l}{\textbf{Encoding Dist.}} & \multicolumn{1}{l}{\textbf{PageRank}} & \multicolumn{1}{l}{\textbf{Betweenness}} & \multicolumn{1}{l}{\textbf{Closeness}} & \multicolumn{1}{l}{\textbf{Degree}} \\ \midrule
    $\alpha = 1$                                        & 0.728                                 & 0.355                                    & 0.246                                  & 0.719                               \\
    $\alpha = 2$                                        & 0.964                                 & 0.760                                    & 0.590                                  & 0.974                               \\
    $\alpha = 3$                                        & 0.889                                 & 0.550                                    & 0.459                                  & 0.872                               \\
    $\alpha = 4$                                        & 0.973                                 & 0.758                                    & 0.598                                  & 0.982                               \\
    \bottomrule
    \end{tabular}
    \label{tab:MiserablesCorrelations}
\end{table}

\end{document}